\documentclass{article}

\usepackage{arxiv}

\usepackage[utf8]{inputenc} 
\usepackage[T1]{fontenc}    
\usepackage{hyperref}       
\usepackage{url}            
\usepackage{booktabs}       
\usepackage{amsfonts}       
\usepackage{nicefrac}       
\usepackage{microtype}      
\usepackage{lipsum}

\usepackage{amsmath}
\usepackage{makeidx}         
\usepackage{graphicx}        
\usepackage{multirow}
\usepackage{multicol}
\title{Analysis of “User-Specific Effect'' and Impact of Operator Skills on Fingerprint PAD Systems}

\author{
  Giulia Orrù \\
  \texttt{giulia.orru@unica.it} \\
   \And
Pierluigi Tuveri \\
  \texttt{pierluigi.tuveri@diee.unica.it} \\\And Luca Ghiani\\
  \texttt{luca.ghiani@diee.unica.it} \\ \And Gian Luca Marcialis\\
  \texttt{marcialis@unica.it} \\ \\
  University of Cagliari - Department of Electrical and Electronic Engineering - Italy 
}
\begin{document}

\maketitle
\begin{abstract}
Fingerprint Liveness detection, or presentation attacks detection (PAD), that is, the ability of detecting if a fingerprint submitted to an electronic capture device is authentic or made up of some artificial materials, boosted the attention of the scientific community and recently machine learning approaches based on deep networks opened novel scenarios.
A significant step ahead was due thanks to the public availability of large sets of data; in particular, the ones released during the International Fingerprint Liveness Detection Competition (LivDet). Among others, the fifth edition carried on in 2017, challenged the participants in two more challenges which were not detailed in the official report. In this paper, we want to extend that report by focusing on them: the first one was aimed at exploring the case in which the PAD is integrated into a fingerprint verification systems, where templates of users are available too and the designer is not constrained to refer only to a generic users population for the PAD settings. The second one faces with the exploitation ability of attackers of the provided fakes, and how this ability impacts on the final performance.
These two challenges together may set at which extent the fingerprint presentation attacks are an actual threat and how to exploit additional information to make the PAD more effective.\end{abstract}


\section{Introduction}
\label{introduction}
In the development of biometric systems, the study of the techniques necessary to preserve the systems' integrity and thus to guarantee their security, is crucial. 
Fingerprint authentication systems are highly vulnerable to artificial reproductions of fingerprint made up of materials such as silicon, gelatine or latex, also called spoofs or fake fingerprints \cite{mats}. To counteract this possibility, Fingerprint Liveness Detection (FLD), also known as Fingerprint Presentation Attaks Detection (FPAD), is a discipline aimed to design pattern recognition-based algorithms for distinguishing between live and fake fingerprints. Detecting presentation attacks is not trivial because we have an arms-race problem, where we potentially are unaware of the materials and methods adopted to attack the system \cite{marasco}.
\newline In the last decade, most of the previously proposed PAD methods have focused on using handcrafted image features, for example LBP, BSIF \cite{bsif}, WLD \cite{wld} or HIG \cite{hig}.
Over the years, PAD methods have been refined and recently CNN-based methods are outnumbering local image descriptors methods \cite{cnn,cnn2}.
\newline In order to assess the performance of FPAD algorithms by using a common experimental protocol and data sets, the international fingerprint liveness detection competition (LivDet) is organized since 2009 \cite{reviewlivdet}. Each edition is characterized by a different set of challenges that must be dealt with by the competitors.
Working on the LivDet competitions since its birth, that is, developing a wide experience in providing large sets of data for assessing the state of the art on this topic, we realized that some data acquisition methods influence the FPAD systems performance.
For this reason, the 2017 edition of the LivDet was focused on assessing how much the composition and the data acquisition methods influence the PAD systems. In particular, we faced here with the followings points: 
\begin{itemize}
    \item the presence of different operators for creating and submitting spoofs, that is, two levels of operators which provided spoofs and attacked the submitted algorithms - can spoofing be an actual threat independently of the ability of attackers, or an expert is necessary to break a fingerprint PAD?; 
    \item  when a PAD must be integrated into a fingerprint verification system, the templates of genuine users are available, and there is no good reason to neglect them even during the PAD design. We referred to the additional information coming from the enrolled user as “user-specific effect'' \cite{userspec}.   
\end{itemize}
These aspects have been analyzed only marginally in the LivDet 2017 report \cite{livdet17} for sake of space. Moreover, the above points were not addressed by the previous editions of LivDet, whose history and results over the years are reported in \cite{reviewlivdet}. 
\newline With regard to the “user-specific effect'', we reported in \cite{userspec} that some specific components or minute details are dependent on the user's skin characteristic. Ref. \cite{userspec} reports a statistical study on the relationships of textural features coming from the same fingerprints of the same user and even from his/her different fingerprints. The correlation we measured suggests that the consequent “bias'' introduced into a PAD can be beneficial if the PAD must be integrated into a fingerprint verification system. Moreover, it does not affect the overall PAD performance when it is used for the generic user of detecting a presentation attack. What we called “user-specific effect'' helps in improving the correct classification of an alive fingerprint of a genuine user, and this is crucial into a fingerprint verification system where the rate of rejected genuine users must be kept as low as possible. From this point of view, the fusion with a non-zero error system, namely, the PAD, may put that rate in danger, as noticed in other publications as \cite{inaction}.
\newline The paper is organized as follows: Section\ \ref{proposed} describes the purpose of the experimentation. The experimental methodology is presented in Section \ref{exprotocol}. Experiments are shown in Section \ref{results}. Conclusions are drawn in Section \ref{conclusion}.
\begin{table}[!h]
\centering
\begin{tabular}[t]{ | l || c|| c | c | c | c | r |}
\hline
\textbf{}	&& \multicolumn{4}{c|}{ \textbf{Accuracy [\%]}}\\ \hline
\textbf{Algorithm}	&\textbf{Type}& \textbf{1} & \textbf{2}	& \textbf{3}	& \textbf{Overall}\\ \hline
SSLFD	&N.A.&93.58&94.33&93.14&93.68\\ \hline
JLW\_A	&Deep Learning&95.08&94.09&\textbf{93.52}&94.23\\ \hline
JLW\_B	&Deep Learning&\textbf{96.44}&\textbf{95.59}&\textbf{93.71}&95.25\\ \hline
OKIBrB20& Hand-crafted	&84.97&83.31&84.00&84.09\\ \hline
OKIBrB30& Hand-crafted	&92.49&89.33&90.64&90.82\\ \hline
ZYL\_1	&Deep Learning&95.91&\textbf{95.13}&91.66&94.23\\ \hline
ZYL\_2	&Deep Learning&\textbf{96.26}&\textbf{94.73}&93.17&94.72\\ \hline
SNOTA2017\_1&N.A.	&95.03&91.26&91.58&92.62\\ \hline
SNOTA2017\_2&N.A.	&94.04&86.72&86.74&89.17\\ \hline
ModuLAB	&Deep Learning&94.25&90.40&90.21&91.62\\ \hline
ganfp	&Deep Learning&95.67&93.66&\textbf{94.16}&94.50\\ \hline
PB\_LivDet\_1	&Hand-crafted&93.85&89.97&91.85&91.89\\ \hline
PB\_LivDet\_2	&Hand-crafted&92.86&90.43&92.60&91.96\\ \hline
hanulj	&Deep Learning&\textbf{97.06}&92.34&92.04&93.81\\ \hline
SpoofWit&Hand-crafted	&93.66&88.82&89.97&90.82\\ \hline
LCPD &Hand-crafted&89.87&88.84&86.87&88.52\\ \hline
PDfV&Hand-crafted&92.86&93.31&N.A.&N.A.\\ \hline
\end{tabular}
\caption{Types and accuracy of the algorithms participating in LivDet 2017. The accuracy of the three sensors used (\textbf{1} = Green Bit, \textbf{2} = Digital Persona, \textbf{3} = Orcanthus) and the overall have been reported.}
\label{tab:overall}
\end{table}
\section{Proposed analysis} 
\label{proposed}

In real applications, the FPAD system works together with a recognition system in order to protect it from spoofing attacks. 
The integration of the PAD to the authentication system may reduce the recognition performance due to some live genuine sample's rejection \cite{inaction}. 
In last years it was noticed that the presence of the same users both in the train set and in the test set, increased the PAD accuracy of the system. 
In fact, the existence of artifacts due to the human skin (person-specific) and to the particular curvature of ridges and valleys (finger-specific) can impact in PAD systems' performance \cite{userspec,userspec2} and can be exploit to improve the integrated system. 
In particular, the data acquired during the recognition system's initial enrollment phase can be used to lower the PAD error rate and consequently, to improve the integrated system performance.
This approach may be referred to as “user-specific", whilst the other one may be referred to as “generic user".
It is important to emphasize that stand-alone PAD systems are usually considered based on “generic user" approaches. Usually, a generic user approach is claimed to avoid biased results. This was evident in some early works \cite{quality,jia} where an unexpected and unexplained drop of error rates occurred when some users were present both in the train and in the test set, because the user-specific effect was not yet clear in the mind of the scholars.
\newline However, this biasing may be desired under certain circumstances. In fact, as already mentioned, the positive influence of the user-specific effect on the live/fake classification can be exploited in real applications, where the enrolment phase allows to record the user fingerprint image previously.
It is therefore possible to state that the “generic-user” approach is the main resource when the final user population is unknown (for example in forensic applications or border checks); the “user-specific" approach can be used to improve security of an integrated fingerprint verification system (for example authentication on mobile or bank account), in which the attacks are finalize to break the verification step and not to simply being disguised for another person.
\newline On the other hand, common sense tells us that the system's performance is influenced by the attacker's ability to replicate the fingerprint.
Although it is known that fakes can be produced with commonly used materials, even if these are of excellent quality, the attacker must know how to use them to spoof the system.
In fact, we believe that an average skilled attacker may have the technique needed to circumvent the matcher, but more skills are needed to spoof an integrated system (both matcher and liveness detector).
Although the impact of image quality has been used to classify fingerprints \cite{quality,quality2}, to the best of our knowledge, no one has ever analyzed the impact of the operator's skill level.
\begin{table*}[t]
\centering
\begin{tabular}{|c|c|c|c|c|c|c|c|c|c|c|c|c|}
\hline
                 & \multicolumn{4}{c|}{\textbf{Train}}      & \multicolumn{4}{c|}{\textbf{Test}}\\ \hline
\textbf{Dataset} & Live & Wood Glue & Ecoflex & Body Double & Live & Gelatine & Latex & Liquid Ecoflex\\ \hline
Green Bit        &1000&400&400&400&1700&680&680&680\\ \hline
Orcanthus        &1000&400&400&400&1700&680&658&680\\ \hline
Digital Persona  &999&400&400&399&1700&679&670&679\\ \hline
\end{tabular}%

\caption{Number of samples for each scanner and each part of the dataset.}
\label{tab:datasetComposition}
\end{table*}

\section{Dataset and experimental protocol}
\label{exprotocol}

LivDet 2017 \cite{livdet17} consisted of data from three fingerprint sensors: Green Bit DactyScan84C, Orcanthus Certis2 Image and Digital Persona U.are.U 5160.
It is composed of almost 6000 images for each scanner.  Live images came from multiple acquisitions of all fingers of different subjects. 
\newline The LivDet 2017 fake images were collected using the cooperative method.  Each dataset consists of two parts, the train set, used to configure the algorithms, and the test set, used to evaluate the algorithms performance.
Moreover, the materials used in the training set are different with respect to the test set as we can see in Table \ref{tab:datasetComposition}.
\newline The test set can be partitioned into three parts, a user-specific (US) portion, composed by fingerprint belonging to users already present in the train set, and the other two partitions are generic user (GU), meaning they are composed by fingerprint belonging to user not present in the train set, in which the fakes are made by two different operators HS (High Skilled) and LS (Low Skilled), described in more detail below. The false samples of the US partition were made by the operator HS.
On the basis of this subdivision we therefore had: 
\begin{equation}
TestSet = \{US (HS), GU (HS), GU (LS)\}
\nonumber\end{equation}
Using the two partitions, $US (HS)$ and $GU (HS)$, separately to test the algorithms, we can compare the user-specific effects with the generic-user case.
In this experiment we left out the $GU (LS)$ partition to avoid an operator bias. 

The second evaluation concerned the comparison between operators with different degrees of skill in replicating fingerprints.
In particular, operator HS  has a high experience in creating fakes because he has participated in multiple LivDet editions, while for operator LS this was the first edition in which he participated. Fingerprint replication requires a lot of experience, as the accuracy of the fakes heavily affects system performance.
For this evaluation, partitions $GU (HS)$ and $GU (LS)$ were used separately as test set. The two operators used the same acquisition protocol using the same materials.

The Table \ref{tab:overall} shows the results of LivDet 2017 for the three sensors, the total accuracy and the type of algorithm presented, making a distinction between deep learning based or hand-crafted features based. The table shows that in 2017 in FPAD the solutions adopted are equally distributed between deep learning and hand-crafted algorithms. Furthermore, we see that the performance of the algorithms that use deep learning techniques are better than the classic ones of about 10 percentage points.
\begin{table}[t]
\renewcommand{\arraystretch}{1.3}
\begin{tabular}{c|c|c|c|c|c|}
\cline{2-6}
\multirow{2}{*}{\textbf{}}                   & \multirow{2}{*}{\textbf{\begin{tabular}[c]{@{}c@{}}Average\\ Accuracy [\%]\end{tabular}}} & \multicolumn{4}{c|}{\textbf{Average EER [\%]}}                                                        \\ \cline{3-6} 
                                             &                                                                                      & \textbf{High skilled} & \textbf{Low skilled} & \textbf{User-specific} & \textbf{Generic User} \\ \hline
\multicolumn{1}{|c|}{\textbf{Deep Learning}} & $94.0(\pm1.9)$                                                                            & $8.1(\pm2.9)$              & $6.5(\pm2.4)$             & $2.6(\pm1.4) $              & $8.1(\pm2.9)$                 \\ \hline
\multicolumn{1}{|c|}{\textbf{Hand-crafted}}  & $90.0(\pm3.1)$                                                                            & $12.5(\pm4.6)$             & $10.8(\pm3.7)$            & $5.7(\pm3.4)$               & $12.5(\pm4.6)$                \\ \hline
\end{tabular}
\caption{Average accuracy and EERs [\%] comparison between deeep learning and hand-crafted  methods. The distinction  between User-Specific and Generic User experiments  and  between  high skilled operator and a low skilled operator is reported. The average accuracy refers to the entire test set.}
\label{tab:deephand}
\end{table}
\section{Results}
\label{results}
In this session the comparisons between User-Specific (US) and Generic User (GU) experiments and between highly skilled operator (HS) and a lowly skilled operator (LS) experiments were reported.

\begin{figure}[h!]
  \centering
    \includegraphics[width=11cm]{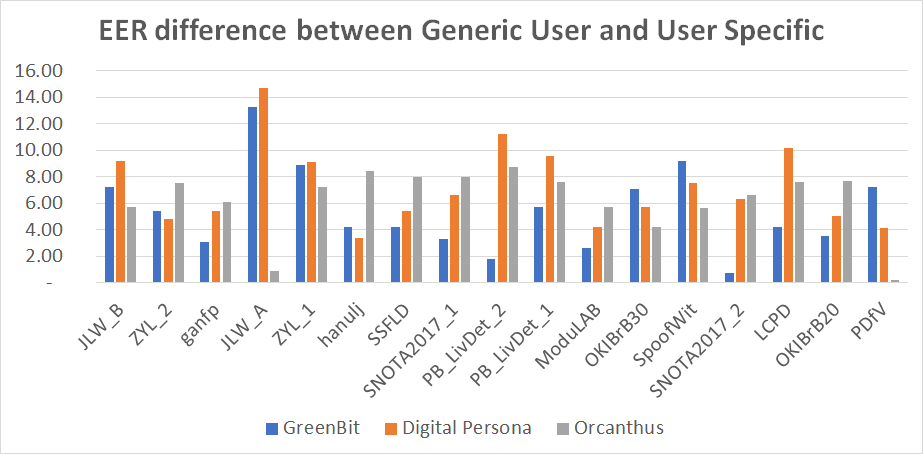}
   \caption{\label{fig:userspec}Comparison between User Specific and Generic User experiments. Each bar is the result of the EER difference between US and GU.}
\end{figure}

In the previous papers the user-specific effect was highlighted only for PAD systems based on hand-crafted features.
The high number of algorithms participating in LivDet 2017 based on deep learning methods allow us to have a clearer view on this effect.
Table \ref{tab:deephand} shows a comparison between deep-learning based methods and handcrafted features based methods, obtained by comparing the average accuracy and the EERs related to the analyzed case studies. 
This table shows that deep learning appears to be more competitive than hand-crafted techniques, as anticipated in Table \ref{tab:overall}. Furthermore it is clear that even the standard deviation is more limited and therefore it is an index of how the deep learning performances have more or less the same performance and with smaller fluctuations with respect to the others.
The table clearly confirms the existence of the user-specific effects for all the algorithms presented: on average the error for the generic user system is more than double the user-specific one.
This result is confirmed by Fig. \ref{fig:userspec}, in which the EER difference between GU and US systems has been reported for each algorithm presented.

\begin{figure}[t]
  \centering
    \includegraphics[width=11cm]{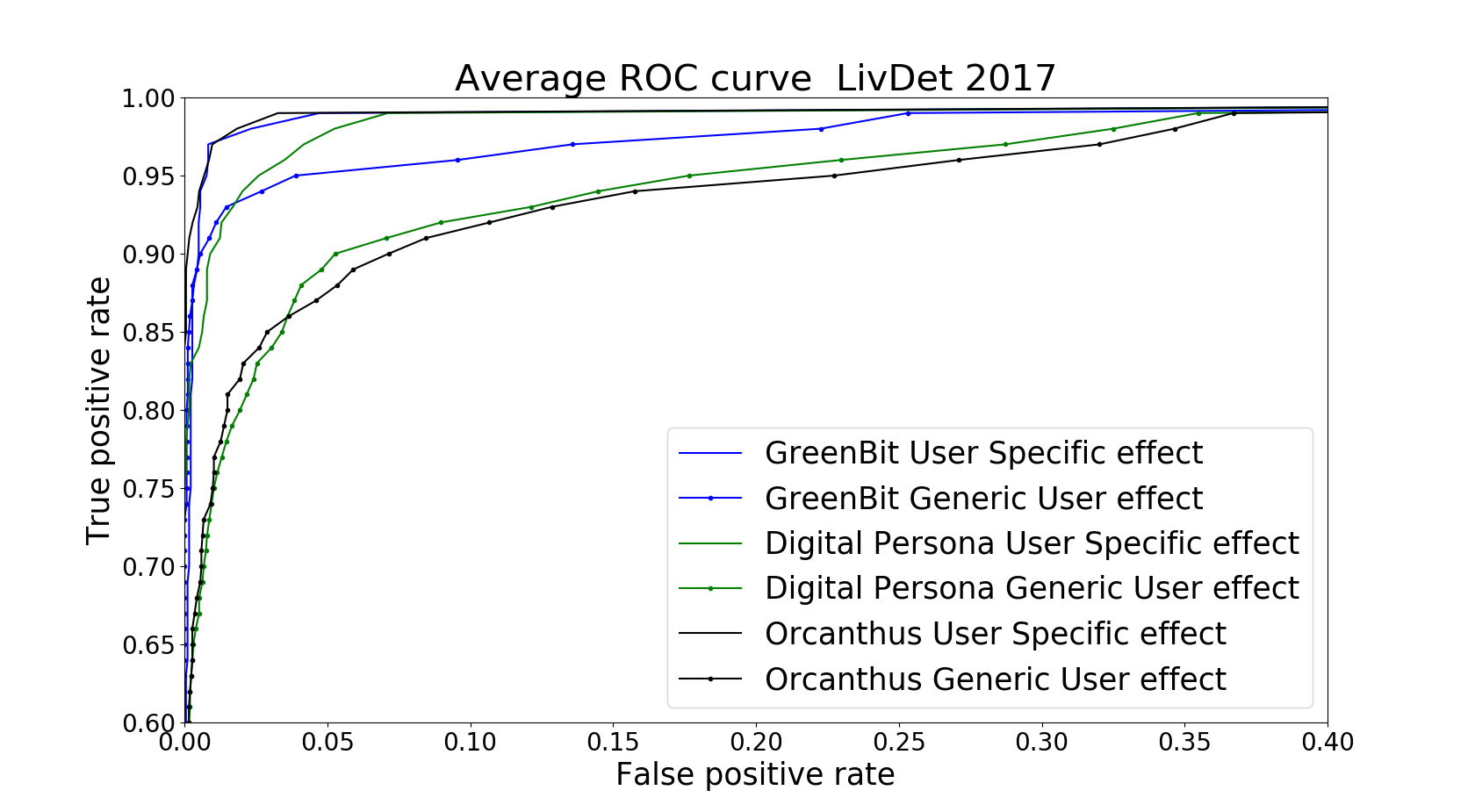}
   \caption{\label{fig:AverageROCUser}Average ROC of the three best LivDet2017 algorithms, calculated for the three scanners. In the ROC the blue, green and black line respectively indicates the Green Bit, Digital Persona and Orcanthus data sets. In the graph the continuous line indicates the US effect, instead the dotted line indicates the trend of the GU ROC.}
\end{figure}

\begin{figure}[t]
  \centering
    \includegraphics[width=11cm]{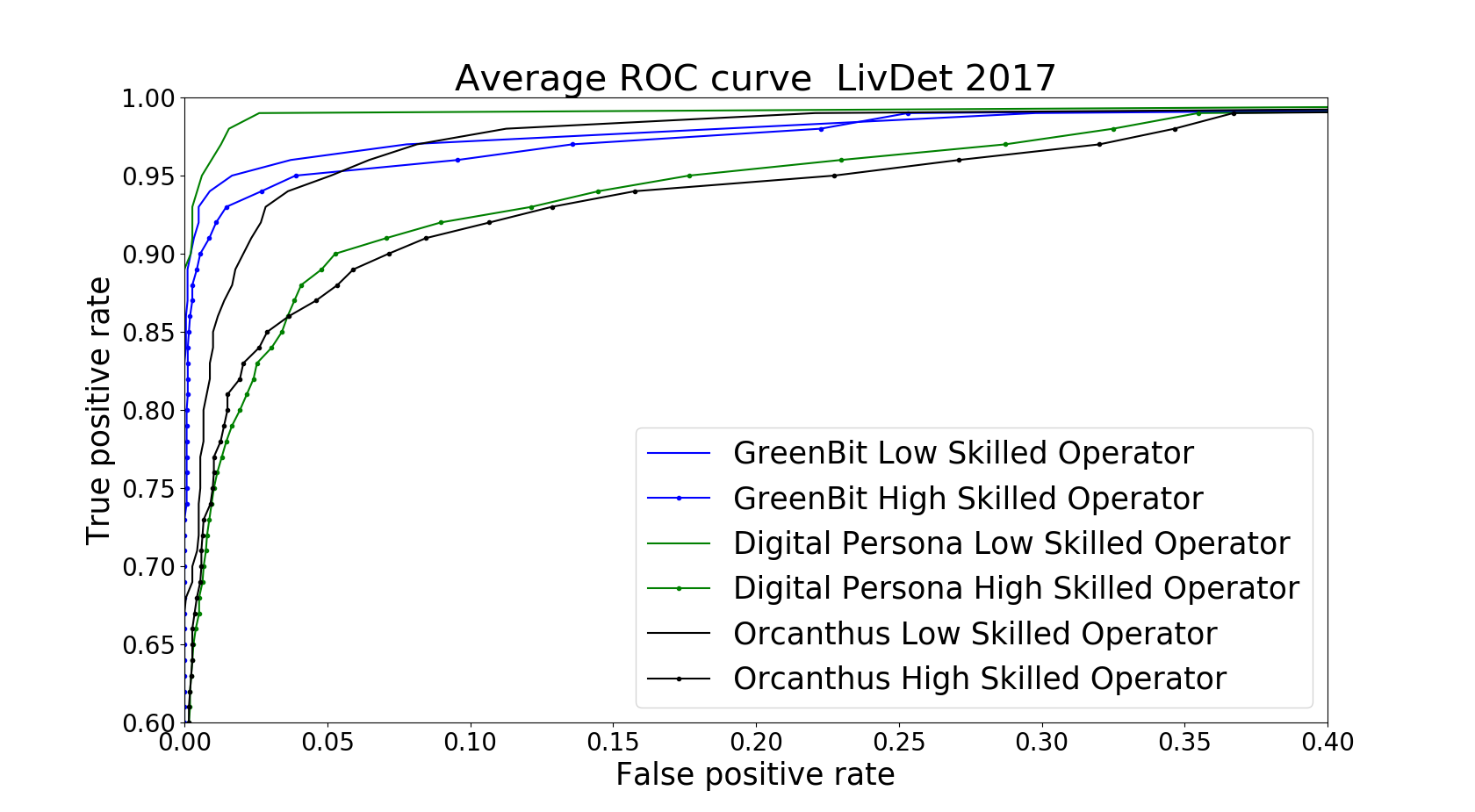}
   \caption{\label{fig:AverageROCOperator}Average ROC of the three best LivDet2017 algorithms, calculated for the three scanners. In the ROC the blue, green and black line  respectively indicates the Green Bit, Digital Persona and Orcanthus data sets. In the graph the continuous line indicates the Low skilled operator in the spoof fabrication fingerprint, instead the dotted line indicates the trend of High skilled operator making fake samples in the ROC curve.}
\end{figure}

To get a graphical view of these results, the ROCs related to the comparisons between user-specific and generic user for the average of the three best algorithms are reported in Fig. \ref{fig:AverageROCUser}. These results demonstrate that the user-specific effect influences liveness detection methods regardless of their characteristics. To assess whether a person-specific or finger-specific effect is more influential, a more in-depth analysis is required. 
This evidence supports the  possible exploitation of user-specific effects to improve the performance of FPAD when integrated with personal identity verification systems, especially when these are used by one or a few users, as in the case of mobile authentication systems.
\begin{figure}[t]
  \centering
    \includegraphics[width=11cm]{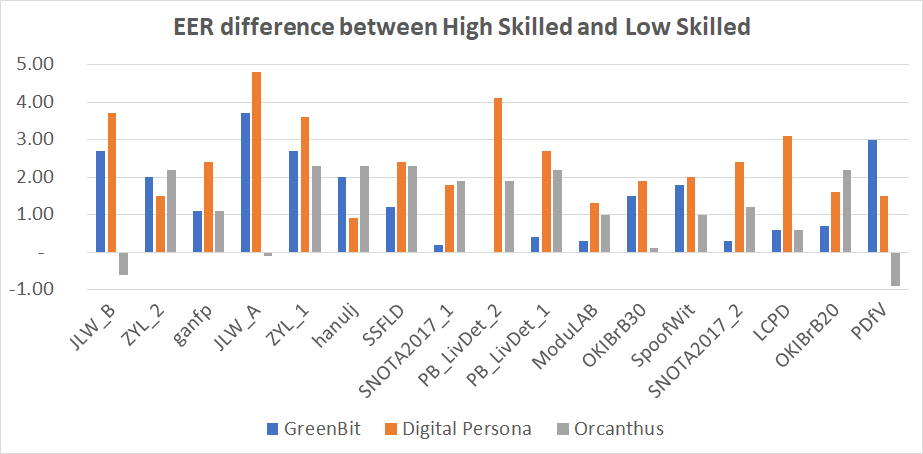}
   \caption{\label{fig:operators}Comparison between high skilled operator and low skilled operator. Each bar is the result of the EER difference between HS and LS.}
\end{figure}
\newline The other analysis carried out concerns the influence of the operator's ability to spoof PAD systems.
The results, shown in Tab. \ref{tab:deephand} and Fig. \ref{fig:operators}-\ref{fig:AverageROCOperator}, confirm the intuitive hypothesis that the attacker's skills profoundly influence the performance of these systems for both hand-crafted and deep learning methods.
In particular, from Fig. \ref{fig:AverageROCOperator}, which shows the comparisons between a low skilled and high skilled operator for the average of the three best algorithms, it is possible to deduce that this difference is not only due to the ability to create fakes, but also to the ability to use fakes to “cheat" the system. In fact, a skilled operator produces excellent fakes and knows how to use them to spoof the FPAD system, thus induces a higher EER. Depending on the sensor, the difference between the continuous curve (LS) and the dotted curve (HS) is more pronounced. Above all, the curves regarding the Green Bit sensor, which is the simplest to use as it has a scanner area almost double compared to the other sensors, have a similar trend. Conversely, the other two sensors, in addition to a smallest area, are much more difficult to use and a correct acquisition require a remarkable experience.
\section{Conclusions}
\label{conclusion}
In this paper, we extended the report on the LivDet 2017 competition edition, in order to point out the main results about two more challenges proposed which were not detailed in the official report for sake of space.

The obtained results can be viewed as a demonstration on what the state of the art on fingerprint PAD can reach on both challenges. Participants used deep learning and hand-crafted features based methods for the task, also the number of competitors was the biggest with respect to other editions.

Reported results allowed us to confirm that the positive influence of the user-specific effect on the liveness classification can be exploited in integrated systems where the enrolment phase is necessary, thus making available samples of the specific user for the PAD design.

Furthermore, it has been shown that the operator's ability to create and use fake replicas heavily affects performance. This pointed out that, to being a serious threat for fingerprint verification system a minimum level of skill is required beside the ability to provide reliable replicas of the targeted user.
%
%
%
%
\begin{small}
%
%

%
%

\end{small}

\end{document}